\documentclass[3p, two-column, final]{elsarticle}

\usepackage[colorlinks=true,urlcolor=black]{hyperref}
\usepackage{amsmath, amssymb}
\usepackage{color}
\usepackage{soul}
\usepackage{subcaption}

\journal{Computerized Medical Imaging and Graphics}

\biboptions{authoryear}

\begin{document}

\begin{frontmatter}

\title{BronchoPose: an analysis of data and model configuration for vision-based bronchoscopy pose estimation}

\author[mainaddress,secondaryaddress]{Juan Borrego-Carazo\corref{correspondingauthor}}
\cortext[correspondingauthor]{Corresponding author. Work performed during a research stage at Computer Vision Center.}
\ead{juan.borrego@uab.cat}

\author[mainaddress]{Carles Sanchez}
\author[secondaryaddress]{David Castells-Rufas}
\author[secondaryaddress]{Jordi Carrabina}
\author[mainaddress]{Débora Gil}

\address[mainaddress]{Computer Vision Center, Universitat Aut\`onoma de Barcelona, 08193 Cerdanyola del Vallès, Spain}
\address[secondaryaddress]{Microelectronics and Electronic Systems Department, Universitat Autònoma de Barcelona, 08193 Cerdanyola del Vallès, Spain}

\begin{abstract}
Vision-based bronchoscopy (VB) models require the registration of the virtual lung model with the frames from the videobronchoscopy to provide an effective guidance during the biopsy. The registration can be achieved by either tracking the position and orientation of the bronchoscopy camera, or by calibrating its deviation from the pose (position and orientation) simulated in the virtual lung model. Recent advances in neural networks and temporal image processing have provided new opportunities for guided bronchoscopy. However, such progress has been hindered by the lack of comparative experimental conditions.

In the present paper we share a novel synthetic dataset allowing for a fair comparison of methods. Moreover, this paper investigates several neural network architectures for the learning of temporal information at different levels of subject personalization. In order to improve orientation measurement, we also present a standartized comparison framework and a novel metric for camera orientation learning.  Results on the dataset show that the proposed metric and architectures, as well as the standardized conditions, provide notable improvements to current state of the art camera pose estimation in videobronchoscopy.  

\end{abstract}

\begin{keyword}
videobronchoscopy guiding, deep learning, architecture optimization, datasets, standardized evaluation framework, pose estimation
\end{keyword}

\end{frontmatter}

\section{Introduction}

Early detection is fundamental for lung cancer mortality reduction \citep{pastorino_prolonged_2019, de_koning_reduced_2020}. After a suspicious pulmonary lesion (PL) has been detected through a computed-tomography (CT) scan, decisive diagnosis can only be achieved through a biopsy. Recent advances in sensorics and imaging have improved the sensitivity yield of navigational bronchoscopy (NB) \citep{asano_virtual_2014, ishiwata_advances_2019}, establishing it as a solid alternative to percutaneous approaches, which have a higher degree of medical complications \citep{han_diagnosis_2018, gould_evaluation_2013}.

Among the different methods for NB, vision-based NB (VNB) stands out for its low cost, accessible configuration, and reliability. In such method, a virtual model of the patient pulmonary system is built from CT scans \citep{mori_improvement_2008, skalski_3d_2010, gil_segmentation_2019} and the optimal path to the PL is defined. The physician should replicate this path during the intervectional bronchoscopy. Therefore, during navigation, VNB models require the registration of the virtual lung model with the frames from the videobronchoscopy to provide an effective guidance during the biopsy. The registration can be achieved by, either tracking the position and orientation of the bronchoscopy camera  \citep{chien_bronchoscope_2020, byrnes_construction_2014, luo_scale_2011, luo_development_2012}, or by callibrating its deviation from the pose (position and orientation) simulated in the virtual lung model.

\begin{table*}[!t]
    \centering
    \resizebox{\textwidth}{!}{
    \begin{tabular}{|c|c|c|c|c|c|c|c|c|}\hline
         Method & Year &   Image Size  & Tracking Type  & PE (mm) & AE (º) & CTF ($\%$) \\
         \hline
         \citep{bricault_registration_1998} & 1998 & 100x100 &  Local/Global & 2 & 5 & - \\
         \hline
         \citep{mori_method_2001} & 2001 &   - & Local & - & - &  79 \\
         \hline
         \citep{helferty_combined_2002} & 2002 & - & Local & - & - & - \\
         \hline
         \citep{mori_tracking_2002} & 2002 & 410x410 &  Local & - & - & 73.37 \\
         \hline
         \citep{deligianni_patient-specific_2004} & 2004 & 454x487 & Local  & 3 $\pm$ 2.26 & 2.18 $\pm$ 1.63 & - \\
         \hline
         \citep{nagao_fast_2004} & 2004 & - & Local & - & - & 77.79 \\
         \hline
         \citep{shinohara_branch_2006} & 2006 &  30x30 & Local & - & - & 76.4 \\
         \hline
         \citep{khare_improved_2009} & 2009 &  - & Local  & - & - & - \\
         \hline
         \citep{khare_toward_2010} & 2010 &  - &  Global & - & - & 89  \\
         \hline         
         \citep{luo_scale_2011} & 2011 &  362x370 & Local & - & - & 83.2 \\
         \hline
         \citep{luo_manismc_2011} & 2011 & 30x30 & Local & - & - & 70.2  \\
         \hline
         \citep{luo_robust_2012} & 2012 &   362x370 &  Local & 3.72 & 10.2 & - \\
         \hline
         \citep{luo_discriminative_2014} & 2014 & 256x263 & Local & 4.5  & 12.3 & - \\
         \hline
         \citep{shen_robust_2015} & 2015 &   487x487 & Local & 8.48 $\pm$ 6.29 & - & - \\
         \hline
         \citep{esteban-lansaque_stable_2016} & 2016 &  - &  Global & - & - & - \\
         \hline
         \citep{visentini-scarzanella_deep_2017} & 2017 &  50x50 & Local & 1.5 & - & - \\ 
         \hline
         \citep{shen_branchbifurcation_2017} & 2017 &   - & Local & 5-15\footnote{Original article do not provide a value but a graphic. Value approximately inferred from it.} & - & - \\
         \hline
         \citep{sganga_offsetnet_2019} & 2019 &  - & Local & 2.4 & 3.4 & 90.2  \\
         \hline\citep{shen_context-aware_2019} & 2019 & 307 × 313 & Local & 3.18 $\pm$ 2.34  & - & - \\
         \hline
         \citep{zhao_generative_2020} & 2020 &   256x256 &  Local & 1.17 & 9.71 & - \\
         \hline
         \citep{wang_visual_2020} & 2020 &   440x440 &  Local & 3.02 & - & 78.1 \\
         \hline
         \citep{banach_visually_2021} & 2021 &  & Local & 6.2 $\pm$ 2.9 & - & -\\
         \hline
    \end{tabular}}
    \caption{Comparison among bronchoscopic tracking studies with regards to data and evaluation characteristics. Notably, none of the methods share a dataset (currently there is no publicly available dataset for this task) or publish their code. Moreover, metrics, although aiming to measure the same quantities, are different or lacking in some cases. Metrics shown are umbrella terms for measuring the position error (PE), angle error (AE) and the number of correctly tracked frames. Tracking type \citep{khare_toward_2010} refers to the type of information provided by the tracking: global type positions the bronchoscope in macro terms, e.g. 3rd bifurcation, while local, gives  information with regards to position and angle of the bronchoscope. 
    }
    \label{tab:comparison}
\end{table*}

Traditionally, the problem of image-based tracking in VNB has been solved through geometric and hand-crafted methods.  Feature generation \citep{chien_bronchoscope_2020, byrnes_construction_2014, sanchez2015navigation, luo_development_2012} and similarity measures \citep{luo_discriminative_2014, shen_robust_2015, luo_manismc_2011, khare_toward_2010} were commonly used for such purpose, accounting, however, with tracking errors and large execution times. 

Recently, supervised data-intensive learning methods, such as neural networks (NNs) \citep{sganga_autonomous_2019, visentini-scarzanella_deep_2017, zhao_generative_2020, shen_context-aware_2019}, have been used for localization and tracking in bronchoscopies, providing better results than previous methods. %and state-of-the-art. 
Moreover, temporal learning techniques have recently been applied to other endoscopic modalites \citep{turan_endo-vmfusenet_2017}, but has not been appropriately tested in bronchoscopy.
Additionally, depth information has lately been extensively used to improve tracking \citep{recasens_endo-depth-and-motion_2021, banach_visually_2021, shen_context-aware_2019, liu_computer_2020}, mixing it with generative neural networks \citep{zhao_generative_2020, shen_context-aware_2019, liu_computer_2020, banach_visually_2021}. 
 
Despite these advances, there is a common obstacle among studies: results are affected by a lack of fair comparability due to the absence of public bronchoscopy datasets and the usage of appropriate metrics as a gold standard,  Additionally, learning methods depend on high data availability, often hindering their application in data scarce environments. Table~\ref{tab:comparison} summarizes such situation from bronchoscopic literature: none of the selected studies details public code or data as to allow for a fair comparison. With regards to metrics, position and angle metrics differ or lack in most studies, making further comparison difficult.

%PE describes the distance (in mm) between the obtained position with respect to the ground truth, while AE describes the difference in rotation with respect to the ground truth. Rotation differences are not easily measured as a single value, thus, different metrics are used but they usually express the difference as a single value in degrees. 
%normalized value  measured  as umbrella terms for various position metrics and, although $L_2$ norm is normally used in both cases, that is not strictly considered. CTF stands for correctly tracked frames, which accounts for the total number of virtual frames matching real video frames, determined by visual inspection as in \citep{mori_method_2001, mori_tracking_2002} or by having a PE and AE error below a threshold \citep{merritt_interactive_2013}. Tracking type \citep{khare_toward_2010} refers to the type of information provided by the tracking: global type positions the bronchoscope in macro terms, e.g. 3rd bifurcation, while local, gives  information with regards to position and angle of the bronchoscope.

\begin{figure*}[t!]
    \centering
    \begin{subfigure}{0.44\textwidth}
        \centering
        \includegraphics[width=0.9\linewidth]{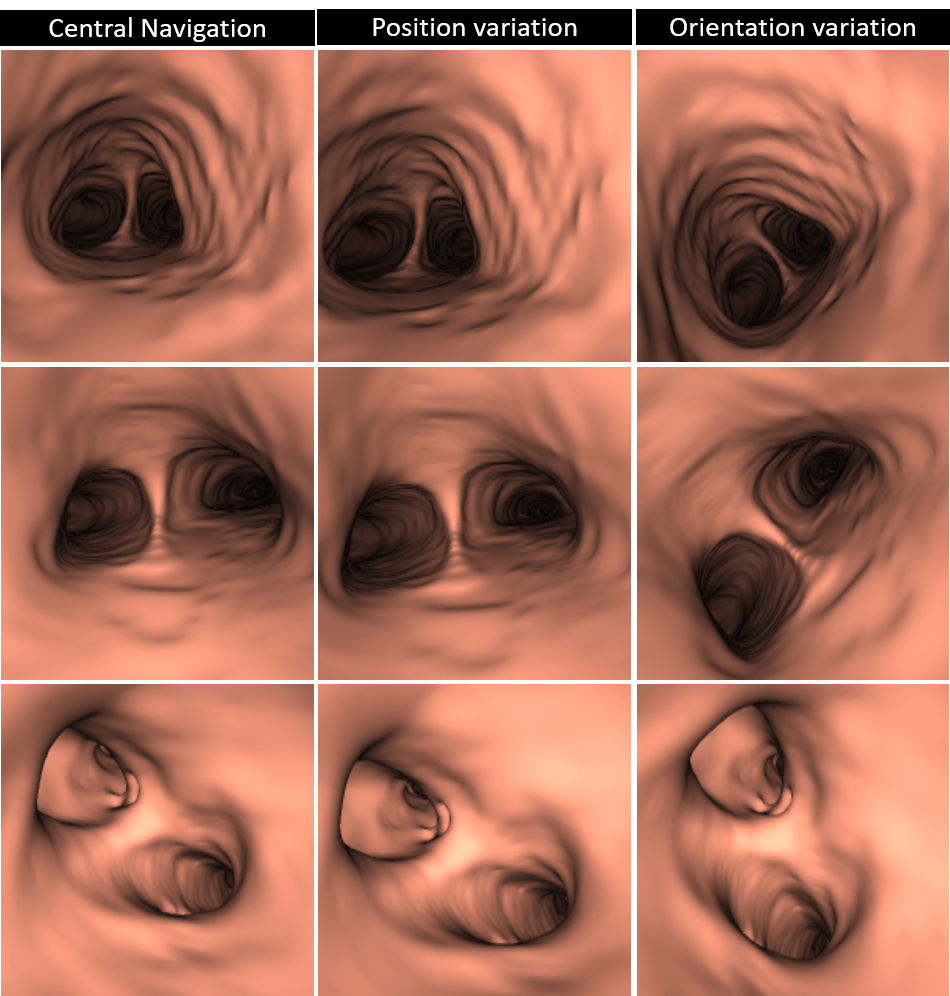}
        \caption{}
        \label{fig:synth}
    \end{subfigure}
    \begin{subfigure}{0.5\textwidth}
        \centering
        \includegraphics[width=0.95\textwidth]{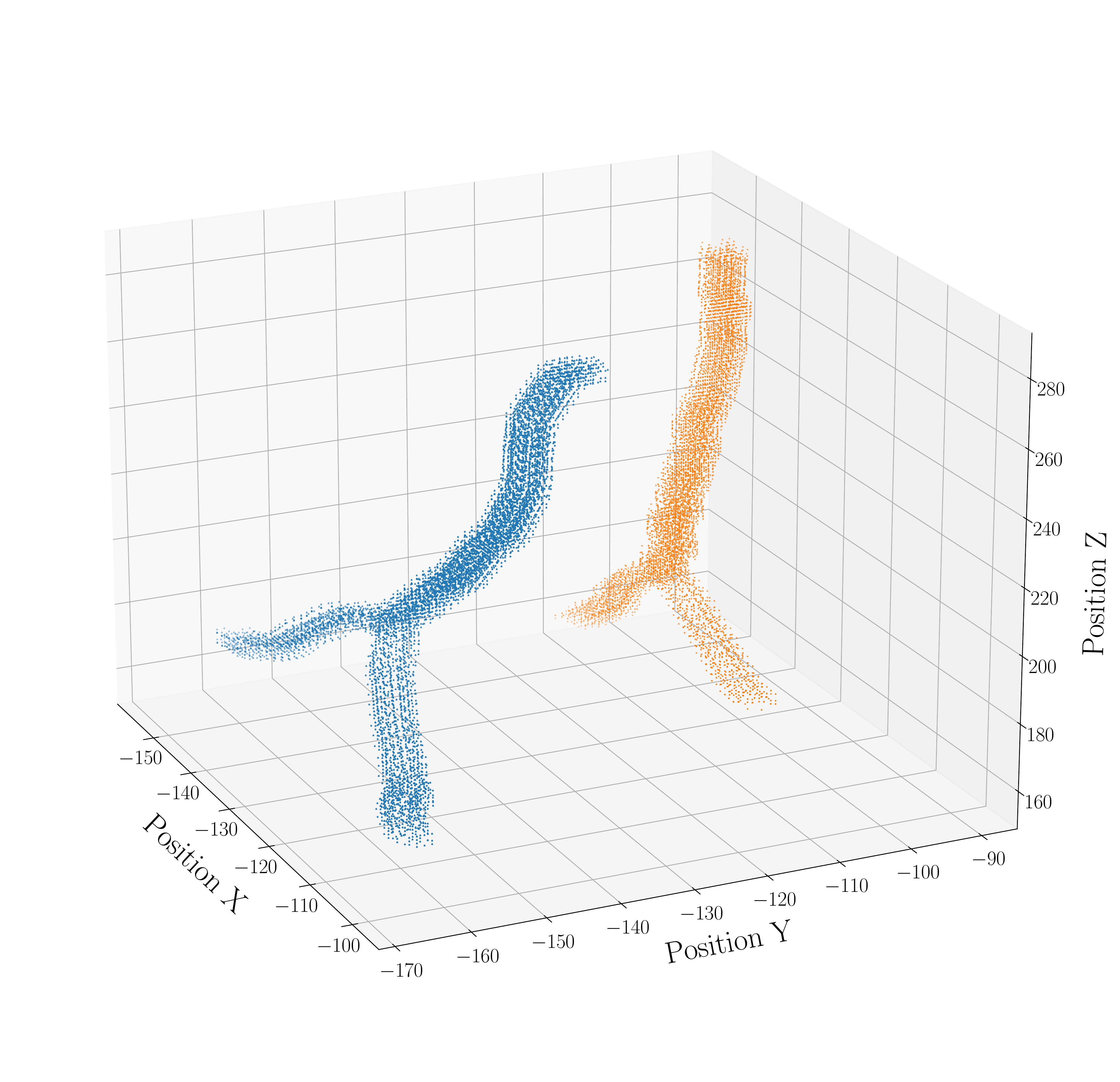}
        \caption{}
        \label{fig:datapath}
    \end{subfigure}
    \caption{Details of the synthetic dataset for bronchoscopy tracking and calibration. \ref{fig:synth}, example of synthetic frames from a trajectory from patient P18 lower left lobe. \ref{fig:datapath}, positions visited by the trajectories corresponding to two different patients: P18 and P20. Both cases only show points pertaining to lower right and left lobe trajectories.}
\end{figure*}

In such a situation, with great progress but also lack of comparability, is of outstanding importance establishing ground points for enabling advances. In the case of a system for pose estimation in intervention guiding, it should address 3 key points: 1) definition of the most appropriate metric and comparison protocol for the evaluation of the estimated pose; 2) determination of the most accurate strategy for the processing of temporal information and 3) the highest generalization level (single or across subject) of models. 

Hence, the goal of this paper is to analyze the performance of different deep learning approaches for image based bronchoscopy tracking using a standardized and fair comparison framework. In particular, we contribute to the following aspects:

\begin{itemize}
    \item {\bf Synthetic Dataset.} We present a bronchoscopy navigation synthetic dataset based on real anatomies to enable fair comparison among methods with a cross-subject setting analysis, as well as, address the data requirements of learning methods.
    
    \item {\bf Evaluation Protocols.} A study and comparison of rotation and position losses and metrics (including a novel one) for bronchoscopy navigation, which helps to establish better grounds for training and evaluation. 
    
    \item {\bf Processing of Temporal Information.} We investigate different solutions for neural network temporal learning. Models such as recurrent NNs (RNNs) \citep{hochreiter_long_1997, schuster_bidirectional_1997}, pseudo-3D convolutions \citep{tran_closer_2018} or 3D convolutions \citep{carreira_quo_2017}, could exploit temporal information in bronchoscopic videos, currently not explored, and provide new results. 

    \item {\bf Population Modelling.} We analyse the different options with regards to data usage for industrial applications: first, with a across patient setting, focused on the development of a general model, and secondly, with an intra-patient setting, by providing a specialized model. 
    
\end{itemize}

%Hence, the second purpose of the present paper is to address such lack of comparability and provide solutions, establishing a base for new advancements. 

%First, to address the lack of public datasets for bronchoscopy, we develop a synthetic VNB dataset that  would allow for fair comparison among methods and also help train learning methods in a scarce data environment such as bronchoscopy.  The dataset will be made publicly available once the paper is published. Moreover, we further analyse the different possibilities with regards to data usage for industrial applications: first, with an outer patient setting, focused on the development of a general model, and secondly, with an intra-patient setting, by providing a specialized model.

%Second, we analyse the current metrics and loss functions used for both training and evaluation of learning methods devoted to VNB, concluding that it has not been fully addressed, with special concern with regards to orientation tracking and the, sometimes inadequate, related metrics \citep{sganga_offsetnet_2019, merritt_interactive_2013}.

%All in all, in the current paper we contribute with three elements all set to enable fair comparison between VNB methods and pursue improvements in their results:

Following sections are organized as follows. Section \ref{sec:dataset} describes the data generation and its characteristics. Section \ref{sec:metrics} presents the different metrics and losses for evaluation. Then, Section~\ref{sec:relativepose}, defines and describes important concepts and composition of the proposed bronchoscopic system and its elements, altogether with a description of the temporal learning architectures and proposed population modelling. Next,  Sections \ref{sec:experimentalsetup} and \ref{sec:results} present the experiments, their setup, and the main results obtained for the system while exploring their significance. Finally, Section \ref{sec:conclusions} concludes the article with the main key points and important takeaways.

\section{Dataset Generation}\label{sec:dataset}

Virtual lung models are built from an own database of computed tomography scans \footnote{\href{http://iam.cvc.uab.es/portfolio/cpap-study-database/}{CVC CPAP Study Database}} \citep{diez-ferrer_positive_2016} using \citep{gil_segmentation_2019} to segment the airways. Virtual airways models are simulated using an own platform developed in C++ and VTK, BronchoX.

From the virtual models, bronchoscope trajectories are simulated from the trachea entrance up until the 4-6 level an covering upper-right, lower-right, upper-left and lower-left lobes. Trajectories are generated from the central navigation path through the luminal central line traversed using the arch-length parameter. Different increments in this parameter allow the simulation of varying velocities across the path. 

For each central path, different variations, both, in position (between $[-2:1:2]$ voxels in each axis) and camera orientation (in the range $[-45:15:45]$ degrees of rotations around the navigation vector) are generated. The variation in camera position implicitly also modifies the camera point of view, since it is given by its position and a point in the central path at a distance $\Delta d$ from the current point. The rotation around this navigation vector introduces a variation in the orientation of the image plane. This way, we simulate a full change in the camera central pose.   

Finally, paths with neighbouring variations are randomly combined along the navigation arc-length parameter in order to simulate realistic trajectories. In total, our dataset has 876 trajectories per patient and lobe, amounting to a total of 842712 frames. 

The dataset has as input values the synthetic frames from the camera view during the trajectory, and as ground truth source values the associated pose inside the VTK airway model coordinate system. The position is in voxel units, and camera view angles are presented in Euler angles. 

Figure~\ref{fig:synth} shows some dataset examples. Each row are different carinas and each column represents variations in position and orientation from the central navigation. Figure~\ref{fig:datapath} shows examples of lower left and right lobes for two different patients. Dataset will be made publicly available upon article acceptance.

\begin{figure*}[!t]
\centering
    \begin{subfigure}{.47\textwidth}
      \centering
      \includegraphics[width=\textwidth, height=0.3\textheight]{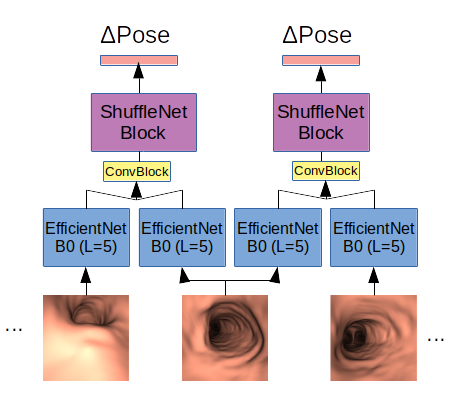}
      \caption{Base network}
      \label{fig:basenetwork}
    \end{subfigure}
    \begin{subfigure}{.47\textwidth}
      \centering
      \includegraphics[width=\textwidth, height=0.3\textheight]{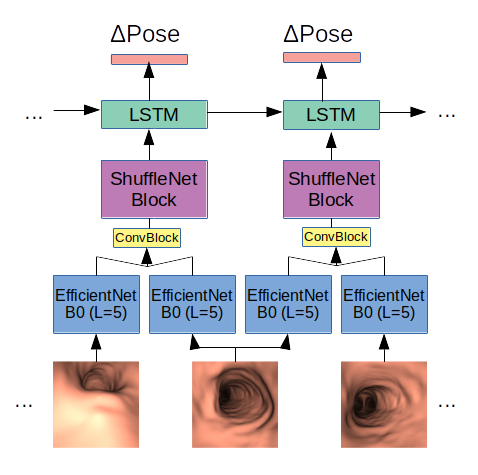}
      \caption{Base network plus recurrence}
      \label{fig:nettemporal}
    \end{subfigure}
    \begin{subfigure}{.47\textwidth}
      \centering
      \includegraphics[width=\textwidth, height=0.3\textheight]{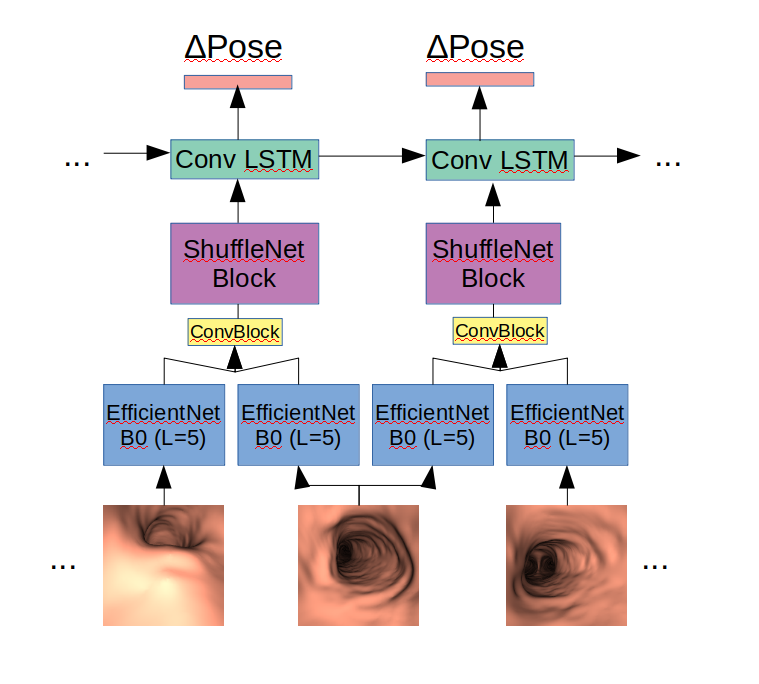}
      \caption{Base network plus convolutional recurrence}
      \label{fig:convtempnetwork}
    \end{subfigure}
    \begin{subfigure}{.47\textwidth}
      \centering
      \includegraphics[width=\textwidth, height=0.3\textheight]{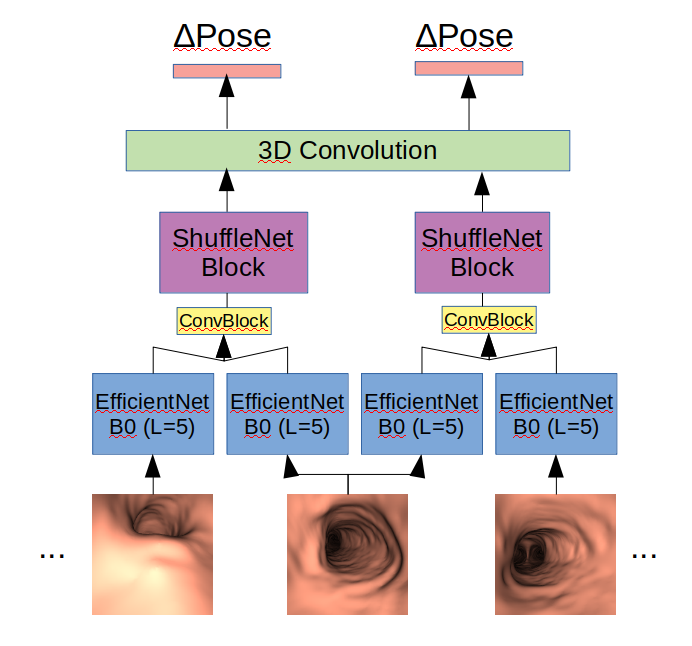}
      \caption{Base network plus 3d convolutions}
      \label{fig:conv3darch}
    \end{subfigure}

\caption{Architectures for bronchoscopy callibration and tracking. (a) is the baseline network without temporal information management, as in \citep{sganga_offsetnet_2019}. (b, c, d) include different mechanisms to manage temporal information across predictions.}
\label{fig:networks}
\end{figure*}

Additionally, the data is processed before training to prepare the inputs and ground truths. For every two pair of images in a sequence, the difference in position and rotation between them is computed. Both components, the difference in position $\Delta{p} =(\Delta x, \Delta y, \Delta z) \in \mathbb{R}^{3}$, and the difference in orientation $\Delta{o}=(\Delta \alpha, \Delta \beta, \Delta \gamma) \in \mathbb{R}^{3}$, define the difference in pose, $\Delta{\textbf{P}} = (\Delta{p}, \Delta{o}) \in \mathbb{R}^{6}$, which constitutes the ground truth of the system. That is, we predict the difference in position and orientation between two images, thus allowing both tracking and calibration. No standardization is applied to the ground truths, while images are standardized through mean subtraction and standard deviation division.

\section{Metrics}\label{sec:metrics}

In order to train and validate a system for pose estimation, we need metrics for assessing the error of the predicted rotation and position. As those are two separate components, we can have different metrics for each of them.

The most common choice are either
the mean squared error (MSE), when the metric is a loss
function, or its equivalent, euclidean norm (L2), when it is an evaluation metric. Although they naturally fit the euclidean space of positions, these functions do not necessarily suit the rotation space. An alternative used in the literature is the \textit{direction error} (DE) \citep{merritt_interactive_2013},

\begin{equation}
    DE = \cos^{-1}(\mathbf{v}_{E}\cdot\mathbf{v}_{GT})
\end{equation}

where $\mathbf{v}_{E}$ and $\mathbf{v}_{GT}$ are, respectively, the estimated and ground truth direction vectors. Such direction vectors are computed from the rotation matrix, $R$, and a unitary direction vector, $\mathbf{u}$, as:

\begin{flalign*}
  &\mathbf{v}_{r}= R\cdot \mathbf{u} =&\\
    &\left[\begin{array}{ccc}
      c_{\beta} c_{\gamma} & -c_{\beta} s_{\gamma} & s_{\beta} \\
      s_{\alpha} s_{\beta} c_{\gamma} + c_{\alpha} s_{\gamma}   & -s_{\alpha} s_{\beta} s_{\gamma} + c_{\alpha} c_{\gamma} & -s_{\alpha} c_{\beta} \\
      -c_{\alpha} s_{\beta} c_{\gamma} + s_{\alpha} s_{\gamma} & c_{\alpha} s_{\beta} s_{\gamma} + s_{\alpha} s_{\gamma} & c_{\alpha} c_{\beta}
    \end{array}\right]
    \begin{bmatrix}
     u_x  \\
      u_y  \\
      u_z 
    \end{bmatrix} &&
\end{flalign*}

\noindent where, $c_{i}$ and $s_{i}$ are, respectively, the cosinus and sinus of angle $i$. A common choice for $\mathbf{u}$ is the look-at vector of the virtual camera, usually given by the x-axis, so that: 
\begin{align*}
    \mathbf{v}_{r} = R\cdot  \begin{bmatrix}
     1  \\
      0  \\
      0 
    \end{bmatrix} =
    \begin{bmatrix}
      c_{\beta} c_{\gamma}  \\
      s_{\alpha} s_{\beta} c_{\gamma} + c_{\alpha} s_{\gamma}  \\
      -c_{\alpha} s_{\beta} c_{\gamma} + s_{\alpha} s_{\gamma}
    \end{bmatrix}
\end{align*}
The main issue with such function is that the choice of \textbf{u} affects the perceived rotations. That is, by selecting an specific \textbf{u}, the system is oblivious to the rotations through that direction.

To remedy such problem, we present an alternative metric, the \textit{cosinus error} (CE):

\begin{align*}
   CE =\frac{1}{3} ( ( 1 - \cos( \Delta \alpha_E - \Delta \alpha_{GT})) + \\ 
  ( 1 - \cos( \Delta \beta_E - \Delta \beta_{GT})) 
  +\\  
  ( 1 - \cos( \Delta \gamma_E - \Delta \gamma_{GT})))
\end{align*}

\noindent where $(\alpha_E,\beta_E,\gamma_E)$, $(\alpha_{GT},\beta_{GT},\gamma_{GT})$ are, respectively the estimated and ground truth Euler angles. 

Importantly, all metrics can be used both as loss for training, as well as metric for evaluation of performance.

\section{Relative Pose Estimation}\label{sec:relativepose}

Given a sequence of $L$ image pairs at time $t$, $(I_1^t,I_2^t) \in \mathcal{D}_1 \times \mathcal{D}_2$, for $\mathcal{D}_1$, $\mathcal{D}_2$ a source and target domains, a pose estimation system can be formulated as a function, $f_{\mathcal{P}}$, predicting the difference in pose between the image pairs:

\begin{equation}\label{eq:basefunction}
    f_{\mathcal{P}}: (I_1^t,I_2^t)_{t=0}^{t=L} \longrightarrow (\Delta \textbf{P}^t)_{t=0}^{t=L}, \quad I_i^t \in \mathcal{D}_i, \Delta P^t\in \mathbb{R}^{6}
\end{equation}

\noindent where each domain $\mathcal{D}_i=\mathbb{R}^{C_i \times H_i \times W_i}$ represents RGB images ($C_i=3$) of size $H_i \times W_i$, and the difference in pose for each time $t$ is a vector $\Delta{P^t}= (\Delta{p^t}, \Delta{o^t}) = (\Delta x^t, \Delta y^t, \Delta z^t, \Delta \alpha^t, \Delta \beta^t, \Delta \gamma^t)$ representing the difference in $(x, y, z)$ position coordinates and $(\alpha, \beta, \gamma)$ Euler angles.

In case $I_1^t$, $I_2^t$ are consecutive frames of the same path, $f_{\mathcal{P}}$ is modelling a tracker, and the change in pose through the sequence would be given by accumulating the differences estimated across the video:
\begin{equation}\label{eq:posesteps}
    \textbf{P}_{L} = \textbf{P}^{0} + \sum_{t=0}^{L}\Delta{\textbf{P}}^{t}
\end{equation}
\noindent where $\textbf{P}^{0}$ corresponds to the initial pose vector, and $\textbf{P}^{L}$ to the pose vector after the accumulated changes of the $L$ pose differences. 

If $I_1^t$, $I_2^t$ correspond to frames of different paths, $f_{\mathcal{P}}$ would be modelling a pose callibration. 

The network loss is given by the addition of the position and rotation metrics:

\begin{equation}
    \mathfrak{L} = \mathfrak{L_p} + \mathfrak{L_o}
\end{equation}
\noindent where $p$ refers to position and $o$ to orientation.  
The position loss $\mathfrak{L_p}$ is given by the MSE error,
while for the orientation loss $\mathfrak{L_o}$ we used the three
metrics (MSE, DE and the proposed CE) described in the previous
section.

In any case (tracking or callibration), if the input sequence has more than two image pairs ($L>0$), there are several ways of processing such temporal information in order to improve the difference in pose estimation. All architectures follow Equation~\ref{eq:posesteps} and their scheme can be found at Figure~\ref{fig:networks}. The different architectures used are: a base network working only between two images, and 3 different ways of incorporating temporal information across frames. Next, configuration details for each architecture are presented, altogether with the population modeling approach.

\subsection{Base Network (Baseline)}

The baseline network is a static estimation of pose differences from a single image pair ($L=0$ in Equation~\ref{eq:basefunction}). Each image is passed through a convolution backbone, specifically an EfficientNet-B0 \citep{tan_efficientnet_2020}. Then both feature maps are concatenated and passed through a convolutional block and a ShuffleNet block \citep{ma_shufflenet_2018}, to obtain a suitable performance/latency trade-off. Finally, the resulting feature map is flattened and passed through a fully connected layer to obtain the difference pose prediction between the two images. An illustration of the overall components of the base network can be found at Figure~\ref{fig:basenetwork}.

\subsection{Recurrence (Baseline + LSTM)}

The previous network does not include temporal management. To be able to include such information, the first modification to the base network consists in the addition of a recurrent LSTM \citep{hochreiter_lstm_1997} (Long Short Term Memory) module after the ShuffleNet block. Such convolutional plues recurrent network type can well exploit temporal information, and thus it has been successfully applied to video tasks, such as object tracking \citep{ning_spatially_2017}, action recognition \citep{ullah_action_2018} or video captioning \citep{jin_recurrent_2019}. 

For every pair of images in the sequence we obtain a group of feature maps. Each one is flattened and a vector, $\mathbf{v}$, of dimension $(L, F)$, is built, where $L$ is the number of image pairs and $F$ the size of the flattened feature maps. Such vector, $\mathbf{v}$, is the input for the LSTM block. At each step, $l$ from the sequence of $L$ image pairs, the output vector from the LSTM cell is forwarded to a fully connected layer, which finally delivers the difference pose vector. An illustration of the mentioned module is found at Figure~\ref{fig:nettemporal}. 

\subsection{Convolutional recurrence (Baseline + ConvRNN)}

In the previous architecture recurrence required flattening the feature maps so they could be fed to the LSTM. Such flattening destroys visual relations present in the feature maps, thus losing information. To avoid such loss, a possibility is to use a convolutional LSTM \citep{shi_convolutional_2015}, where vectorial operations are substituted by convolution ones. In such way, we are able to maintain the visual structure during recurrence. Flattening, however, is still needed to produce the pose prediction through a fully connected layer after Conv-LSTM. In Figure~\ref{fig:convtempnetwork}, an illustration of the overall structure is presented.

\subsection{3D Convolution (Baseline + 3D)}

An alternative to recurrence for managing temporal information is 3D convolutions \citep{ji_3d_2018, guo_deep_2019}. Once all the feature representations from all image pairs in the sequence are generated, a 4D tensor of size $(L, C, H, W)$   can be built. 

Such tensor is fed to a block of two (Conv3D, BatchNorm, ReLU) layers.  Working at once with the feature maps coming from all image pairs allows to learn relations among them, producing improvements in angle and position estimation. After the 3D convolution block, the result is flattened and fed to a fully connected layer to predict the final difference pose prediction. In Figure~\ref{fig:conv3darch} an illustration of the overall network can be seen.

\subsection{Approaches for Population}\label{subsec:population}

Once the data has been prepared as specified in Section~\ref{sec:dataset}, we define two different approaches to prepare our training and validation scheme. The purpose is to establish the attainable degree of generalization of the models developed in two different industrial settings.

First, the population or cross-subject setting, in which a patient is selected as validation and the rest of patients are used for training. This case, in an industrial environment, would correspond to building a general model with different patients and expect enough generalization of the model as to apply it directly to an unseen patient.

And second, the personalized setting, where validation is performed over the same patients as training, but with different sequences. Such procedure, in an industrial environment, implies that every time we include a patient, models should be retrained. Hence, although the procedure would be more cumbersome, less generalization effort is expected from models.

\begin{table*}
    \centering
    \begin{tabular}{c|c|c|c|c|}
        \cline{2-5}
        Baseline & \textbf{Position Error} & \multicolumn{3}{|c|}{\textbf{Rotation Error}} \\
        \hline
       \multicolumn{1}{|c|}{\textbf{Loss}} & $L_2$ & $L_2$ & DE & CE \\

        \multicolumn{1}{|c|}{$\mathfrak{L_p}_{MSE}$ + $\mathfrak{L_o}_{MSE}$}   &  9.657  $\pm$  5.777  & \textcolor{red}{\textbf{19.037 $\pm$ 51.06}} & 1.608 $\pm$ 0.658 & \textbf{0.803 $\pm$ 0.357} \\

        \multicolumn{1}{|c|}{$\mathfrak{L_p}_{MSE}$ + $\mathfrak{L_o}_{DE}$} & 8.426 $\pm$ 4.569 &  45.196 $\pm$ 46.648 & 1.566 $\pm$ 0.63  & 1.018 $\pm$ 0.411 \\
  
            \multicolumn{1}{|c|}{$\mathfrak{L_p}_{MSE}$ + $\mathfrak{L_o}_{CE}$} & \textbf{7.138 $\pm$ 4.547} & 35.029 $\pm$ 47.231 & \textcolor{red}{\textbf{1.477 $\pm$ 0.65}}  & 0.925 $\pm$ 0.937 \\
        \hline
        \cline{2-5}
        Baseline + LSTM &  \multicolumn{4}{|c|}{} \\
          \hline
       \multicolumn{1}{|c|}{\textbf{Loss}} & $L_2$ & $L_2$ & DE & CE \\
       
        \multicolumn{1}{|c|}{$\mathfrak{L_p}_{MSE}$ + $\mathfrak{L_o}_{MSE}$}   &  4.962  $\pm$ 2.739 & \textbf{19.779 $\pm$ 50.898} & 1.733 $\pm$ 0.567 & \textbf{0.775 $\pm$ 0.362} \\
        
        \multicolumn{1}{|c|}{$\mathfrak{L_p}_{MSE}$ + $\mathfrak{L_o}_{DE}$} & 6.714 $\pm$ 3.495 &  45.229 $\pm$ 47.895 & \textbf{1.528 $\pm$ 0.565}  & 1.020 $\pm$ 0.424 \\
        
            \multicolumn{1}{|c|}{$\mathfrak{L_p}_{MSE}$ + $\mathfrak{L_o}_{CE}$} & \textcolor{blue}{\textbf{4.515 $\pm$ 3.636}} & 36.255 $\pm$ 45.839 & 1.670 $\pm$ 0.718  & 0.791 $\pm$ 0.411 \\
        \hline
        \cline{2-5}
        Baseline + 3D & \multicolumn{4}{|c|}{} \\
          \hline
       \multicolumn{1}{|c|}{\textbf{Loss}} & $L_2$ & $L_2$ & DE & CE \\

        \multicolumn{1}{|c|}{$\mathfrak{L_p}_{MSE}$ + $\mathfrak{L_o}_{MSE}$}   &  7.798  $\pm$ 4.176 & \textcolor{blue}{\textbf{19.499 $\pm$ 52.023}} & \textbf{1.502 $\pm$ 0.726} & \textcolor{blue}{\textbf{0.740 $\pm$ 0.359}} \\

        \multicolumn{1}{|c|}{$\mathfrak{L_p}_{MSE}$ + $\mathfrak{L_o}_{DE}$} & 8.487 $\pm$ 4.757 &  40.985 $\pm$ 45.632 & 1.572 $\pm$ 0.683  & 1.020 $\pm$ 0.411 \\

            \multicolumn{1}{|c|}{$\mathfrak{L_p}_{MSE}$ + $\mathfrak{L_o}_{CE}$} & \textbf{5.068 $\pm$ 2.869} & 36.445 $\pm$ 48.947 & 1.678 $\pm$ 0.534  & 0.7595 $\pm$ 0.356 \\
        \hline
        \cline{2-5}
        Baseline + ConvRNN & \multicolumn{4}{|c|}{} \\
          \hline
       \multicolumn{1}{|c|}{\textbf{Loss}} & $L_2$ & $L_2$ & DE & CE \\

        \multicolumn{1}{|c|}{$\mathfrak{L_p}_{MSE}$ + $\mathfrak{L_o}_{MSE}$}   & 4.550  $\pm$ 3.352 & \textbf{21.163 $\pm$ 50.795} & 1.609 $\pm$ 0.665 & \textcolor{red}{\textbf{0.669 $\pm$ 0.348}} \\

        \multicolumn{1}{|c|}{$\mathfrak{L_p}_{MSE}$ + $\mathfrak{L_o}_{DE}$} & 6.088 $\pm$ 4.686 &  42.226 $\pm$ 46.591 & 1.643 $\pm$ 0.642  & 0.963 $\pm$ 0.441 \\

            \multicolumn{1}{|c|}{$\mathfrak{L_p}_{MSE}$ + $\mathfrak{L_o}_{CE}$} & \textcolor{red}{\textbf{4.487 $\pm$ 3.945}} & 34.881 $\pm$ 48.553 & \textcolor{blue}{\textbf{1.487 $\pm$ 0.663}}  & 0.809 $\pm$ 0.394 \\
        \hline
    \end{tabular}
    \caption{Personalized data scheme. Results for the different loss  and architecture combinations. Values show mean and standard deviation computed among all the paths in the validation set. Results are computed after accumulating tracking through the whole validation paths and only by evaluating last position and orientation, as indicated in Equation~\ref{eq:posesteps}. Bold font indicates best value inside group of row and column. Red indicates best of column, and blue second best of column.}
    \label{tab:lossresultsacum}
\end{table*}
%%%%%%%%%%%%%%%%%%%%%%%%%%%%%%%%%%%%%%%%%%%‰%%%%%%%%%%%%%%%%%%%%%%%%%%%%%%%%%%%%%%%%%%%%%%%%%%
%%%%%%%%%%%%%%%%%%%%%%%%%%%%%%%%%%%%%%%%%%%‰%%%%%%%%%%%%%%%%%%%%%%%%%%%%%%%%%%%%%%%%%%%%%%%%%%
%%%%%%%%%%%%%%%%%%%%%%%%%%%%%%%%%%%%%%%%%%%‰%%%%%%%%%%%%%%%%%%%%%%%%%%%%%%%%%%%%%%%%%%%%%%%%%%
%%%%%%%%%%%%%%%%%%%%%%%%%%%%%%%%%%%%%%%%%%%‰%%%%%%%%%%%%%%%%%%%%%%%%%%%%%%%%%%%%%%%%%%%%%%%%%%

%\begin{figure}[!t]
%        \centering
%        \includegraphics[width=0.45\textwidth]{images/data_increase.pdf}
%        \caption{Effect of data increase in an outer-patient (red) and in an intra-patient setting (green). Average Position Error is represented by a dashed line and Average Cosinus Error is represented by a solid line. The X axis represents the number of sequences per lobe and patient available.}
        \label{fig:intraouter}
%\end{figure}

\begin{table*}
    \centering
    \begin{tabular}{c|c|c|c|c|}
        \cline{2-5}
        Baseline & \textbf{Position Error} & \multicolumn{3}{|c|}{\textbf{Rotation Error}} \\
        \hline
       \multicolumn{1}{|c|}{\textbf{Loss}} & $L_2$ & $L_2$ & DE & CE \\

        \multicolumn{1}{|c|}{$\mathfrak{L_p}_{MSE}$ + $\mathfrak{L_o}_{MSE}$}   &  0.447  $\pm$  0.276  & \textbf{1.142 $\pm$ 8.339} & \textbf{0.649 $\pm$ 0.513} & \textbf{0.147 $\pm$ 0.201} \\

        \multicolumn{1}{|c|}{$\mathfrak{L_p}_{MSE}$ + $\mathfrak{L_o}_{DE}$} & 0.459 $\pm$ 0.282 &  2.529$\pm$ 8.465 & 0.830 $\pm$ 0.590  & 0.472 $\pm$ 0.392 \\
  
            \multicolumn{1}{|c|}{$\mathfrak{L_p}_{MSE}$ + $\mathfrak{L_o}_{CE}$} & \textbf{0.395 $\pm$  0.268} & 1.749 $\pm$ 8.508 & 0.735 $\pm$ 0.571  & 0.188 $\pm$ 0.237 \\
        \hline
        \cline{2-5}
        Baseline + LSTM &  \multicolumn{4}{|c|}{} \\
          \hline
       \multicolumn{1}{|c|}{\textbf{Loss}} & $L_2$ & $L_2$ & DE & CE \\
       
        \multicolumn{1}{|c|}{$\mathfrak{L_p}_{MSE}$ + $\mathfrak{L_o}_{MSE}$}   &  \textcolor{blue}{\textbf{0.353  $\pm$ 0.221}} & \textcolor{blue}{\textbf{1.092 $\pm$ 8.349}} & \textcolor{blue}{\textbf{0.611 $\pm$ 0.501}} & \textcolor{blue}{\textbf{0.136 $\pm$ 0.195}} \\
        
        \multicolumn{1}{|c|}{$\mathfrak{L_p}_{MSE}$ + $\mathfrak{L_o}_{DE}$} & 0.421 $\pm$ 0.241 &  2.516 $\pm$ 8.488 & 0.816 $\pm$ 0.602  & 0.446 $\pm$ 0.371 \\
        
            \multicolumn{1}{|c|}{$\mathfrak{L_p}_{MSE}$ + $\mathfrak{L_o}_{CE}$} & 0.371 $\pm$ 0.240 & 1.888 $\pm$ 8.157 & 0.790 $\pm$ 0.599  & 0.213 $\pm$ 0.250 \\
        \hline
        \cline{2-5}
        Baseline + 3D & \multicolumn{4}{|c|}{} \\
          \hline
       \multicolumn{1}{|c|}{\textbf{Loss}} & $L_2$ & $L_2$ & DE & CE \\

        \multicolumn{1}{|c|}{$\mathfrak{L}_{MSE}$ + $\mathfrak{L}_{MSE}$}   &  0.443  $\pm$ 0.295 & \textcolor{red}{\textbf{0.916 $\pm$ 8.341}} & \textcolor{red}{\textbf{ 0.503 $\pm$ 0.416}} & \textcolor{red}{\textbf{0.093 $\pm$ 0.149}} \\

        \multicolumn{1}{|c|}{$\mathfrak{L}_{MSE}$ + $\mathfrak{L}_{DE}$} & 0.467 $\pm$ 0.289 &  2.293 $\pm$ 8.841 & 0.880 $\pm$ 0.590  & 0.386 $\pm$ 0.327 \\

            \multicolumn{1}{|c|}{$\mathfrak{L}_{MSE}$ + $\mathfrak{L}_{CE}$} & \textbf{0.397 $\pm$ 0.252} & 1.592 $\pm$8.508 & 0.724 $\pm$ 0.559  & 0.167 $\pm$ 0.221 \\
        \hline
        \cline{2-5}
        Baseline + ConvRNN & \multicolumn{4}{|c|}{} \\
          \hline
       \multicolumn{1}{|c|}{\textbf{Loss}} & $L_2$ & $L_2$ & DE & CE \\

        \multicolumn{1}{|c|}{$\mathfrak{L}_{MSE}$ + $\mathfrak{L}_{MSE}$}   & 0.368  $\pm$ 0.245 & \textbf{1.165 $\pm$ 8.348} & \textbf{0.655 $\pm$ 0.502} & \textbf{0.153 $\pm$ 0.197} \\

        \multicolumn{1}{|c|}{$\mathfrak{L_p}_{MSE}$ + $\mathfrak{L_o}_{DE}$} & 0.432 $\pm$ 0.260 & 2.480 $\pm$ 8.449 & 0.888 $\pm$ 0.635  & 0.452 $\pm$ 0.361 \\

            \multicolumn{1}{|c|}{$\mathfrak{L}_{MSE}$ + $\mathfrak{L}_{CE}$} & \textcolor{red}{\textbf{0.355 $\pm$ 0.265}} & 1.80 $\pm$ 8.522 & 0.732 $\pm$ 0.537  & 0.195 $\pm$ 0.242 \\
        \hline
    \end{tabular}
    \caption{Personalized data scheme. Results for the different loss  and architecture combinations evaluated at every image pair. Values show mean and standard deviation computed among all the pairs of images in the paths of the validation set. Bold font indicates best value inside group of row and column. Red indicates best of column, and blue second best of column.}
    \label{tab:lossresults}
\end{table*}

\iffalse
\begin{table*}[!t]
    \centering
    \resizebox{\textwidth}{!}{%
    \begin{tabular}{|c|c|c|c|c|c|}
        \hline
        Config. & Param. (M) & Latency (ms) & GMAC & $L_2$ (voxels) & CE\\
        \hline
         Baseline & \textbf{13.967} & \textbf{28 $\pm$ 2}  & \textbf{1.086} & 0.395 $\pm$ 0.268 &  0.188 $\pm$ 0.237 \\
         
         \hline
         Baseline + LSTM & 14.164  & 28 $\pm$ 1   & 1.087 & 0.371 $\pm$ 0.240 & 0.213 $\pm$ 0.250  \\
         \hline
         Baseline + ConvRNN & 14.142  & 32 $\pm$ 5 & 1.089 & \textbf{0.355 $\pm$ 0.252} & 0.195 $\pm$ 0.242 \\
         \hline
         Baseline + 3D & 13.983  & 40 $\pm$ 13 & 1.223 & 0.397 $\pm$ 0.252 & \textbf{0.167 $\pm$ 0.221} \\
         \hline
    \end{tabular}}
    \caption{Ablation results for the different architectures proposed in Section~\ref{subsec:architecture}. Latency is computed with a NVIDIA RTX 2080Ti. {\bf AQUESTA TAULA JO LA RESERVARIA PEL PAPER ACCELERACIO} \hl{dcr:tb em sembla bé}}
    \label{tab:archablations}
\end{table*}
\fi

\section{Experimental setup}\label{sec:experimentalsetup}

In this section we present the experimental details and the proposed experiments using the different configurations stated in Section~\ref{sec:relativepose}.

Three different experiments are defined:
\begin{enumerate}

\item {\bf Model Optimization.} Loss and architecture combinations are trained and validated on a personalized data setting. 

%A training and selection of personalized models, which consists in a leave-one-out validation on a training set of patient paths to select the best model.  {\bf 4 taules com la taula 2. Una per a cada arquitectura provada.}

\item {\bf Model generalization}. Best architecture and loss combinations from Experiment 1 are selected. They are trained and tested on a population data scheme with a leave-one-out validation structure.
% {\bf 1 taula com la taula 2 per l'arquitectura triada però testejada amb un leave-one-out sobre pacient.} A testing and assessment of models reproducibility, which is a validation of the best model on an independent set of test patients to assess the reproducibility of results.

\item {\bf Model comparison with SoA}. To assess the advantages of the proposed strategy, the best model selected in the first and second experiments is compared to a SoA method.
\end{enumerate}

For each experiment we have selected different groups of data. Specifically, in the case of personalized data scheme, we have selected 15 trajectories per patient and lobe for training, reserving a total of 3 for validation. Each sequence has been divided into a set of pairs of images. In the case of temporal information experiments, we have selected the same number of sequences, but sequences have been split into chunks of 10 image pairs. Additionally, in the population data scheme, we have selected 15 trajectories per patient and lobe for 4 patients for training, and 15 trajectories of 1 patient for validation.

Networks have been trained using two NVIDIA RTX 2080ti, with Pytorch Lightning as a framework \citep{falcon_pytorch_2019}. Adam optimizer with a learning rate of $1e^{-4}$ has been used  and early stopping based on validation loss evolution has been applied. Dropout has been added to the ShuffleNet blocks to avoid overfitting. In the case of temporal learning, networks have been trained using truncated back propagation through time (T-BPTT). A batch size of 512 has been used in all the experiments. With regards to evaluation metrics, the same functions used as losses are used for evaluation. Namely, $L_2$ for position, and $L_2$, DE, and CE, for orientation.

%%%%%%%%%%%%%%%%%%%%%%%%%%%%%%%%%%%%%%%%%%%‰%%%%%%%%%%%%%%%%%%%%%%%%%%%%%%%%%%%%%%%%%%%%%%%%%%
%%%%%%%%%%%%%%%%%%%%%%%%%%%%%%%%%%%%%%%%%%%‰%%%%%%%%%%%%%%%%%%%%%%%%%%%%%%%%%%%%%%%%%%%%%%%%%%
%%%%%%%%%%%%%%%%%%%%%%%%%%%%%%%%%%%%%%%%%%%‰%%%%%%%%%%%%%%%%%%%%%%%%%%%%%%%%%%%%%%%%%%%%%%%%%%
%%%%%%%%%%%%%%%%%%%%%%%%%%%%%%%%%%%%%%%%%%%‰%%%%%%%%%%%%%%%%%%%%%%%%%%%%%%%%%%%%%%%%%%%%%%

\begin{table*}[!t]
    \centering
    \resizebox{\textwidth}{!}{%
    \begin{tabular}{|c|c|c|c|c|c|}
        \hline
        \multicolumn{2}{|c|}{}& Position Error & \multicolumn{3}{|c|}{\textbf{Rotation Error}}\\
        \hline
        Network & Loss & $L_2$ & $L_2$ & DE & CE \\
        \hline
         Baseline + LSTM & $\mathfrak{L}_{MSE}$ + $\mathfrak{L}_{MSE}$ & \textcolor{blue}{16.398 $\pm$ 7.215}  & \textcolor{blue}{29.683 $\pm$ 50.605} & 1.659 $\pm$ 0.691 &  0.863 $\pm$ 0.391 \\
         
         \hline
         Baseline + 3D & $\mathfrak{L}_{MSE}$ + $\mathfrak{L}_{MSE}$  & 18.245 $\pm$ 8.158   & \textcolor{red}{24.104 $\pm$ 52.498} & \textcolor{blue}{1.583 $\pm$ 0.734} & \textcolor{red}{0.717 $\pm$ 0.355}  \\
         \hline
         Baseline + ConvRNN & $\mathfrak{L}_{MSE}$ + $\mathfrak{L}_{CE}$  & \textcolor{red}{14.941 $\pm$ 7.886} & 37.561 $\pm$ 48.519 & \textcolor{red}{1.580 $\pm$ 0.666} & \textcolor{blue}{0.848 $\pm$ 0.388} \\
         \hline

    \end{tabular}}
    \caption{Cross-subject data scheme. Results for loss and architecture combinations that obtained best results in the personalized data scheme.  Results are computed after accumulating tracking through the whole validation path and only by evaluating last position and orientation, as indicated in Equation~\ref{eq:posesteps}. Values show mean and standard deviation computed among all the paths in the validation set. Red indicates best value in column, and blue column's second best. }
    \label{tab:archablationsacum}
\end{table*}

\begin{table*}[!t]
    \centering
    \resizebox{\textwidth}{!}{%
    \begin{tabular}{|c|c|c|c|c|c|}
        \hline
        \multicolumn{2}{|c|}{}& Position Error & \multicolumn{3}{|c|}{\textbf{Rotation Error}}\\
        \hline
        Network & Loss & $L_2$ & $L_2$ & DE & CE \\
        \hline
         Baseline + LSTM & $\mathfrak{L_p}_{MSE}$ + $\mathfrak{L_o}_{MSE}$ & \textcolor{blue}{0.680 $\pm$ 0.319}  & \textcolor{blue}{1.901 $\pm$ 9.913} & \textcolor{blue}{1.017 $\pm$ 0.659} &  \textcolor{blue}{0.324 $\pm$ 0.279} \\
         
         \hline
         Baseline + 3D & $\mathfrak{L_p}_{MSE}$ + $\mathfrak{L_o}_{MSE}$  & 0.711 $\pm$ 0.339   & \textcolor{red}{1.314 $\pm$ 9.905} & \textcolor{red}{0.739 $\pm$ 0.547} & \textcolor{red}{0.172 $\pm$ 0.203}  \\
         \hline
         Baseline + ConvRNN & $\mathfrak{L_p}_{MSE}$ + $\mathfrak{L_o}_{CE}$  & \textcolor{red}{0.678 $\pm$ 0.334} & 2.348 $\pm$ 9.965 & 1.133 $\pm$ 0.712 & 0.392 $\pm$ 0.305 \\
         \hline

    \end{tabular}}
    \caption{Cross-subject data scheme. Results for loss and architecture combinations that obtained best results in the personalized data scheme.  Results are computed for every pair of images. Values show mean and standard deviation computed among all the pairs of images in the paths of the validation set. Red indicates best of column, and blue column's second best. }
    \label{tab:archablations}
\end{table*}

\section{Results and discussion}\label{sec:results}

In the next section we present the results corresponding to the three experiments stated in Section~\ref{sec:experimentalsetup}. Code used to build the system and reproduce experiment results will be published upon acceptance. 

\subsection{Model Optimization}

Table~\ref{tab:lossresultsacum} shows results (mean $\pm$ standard deviation) in the validation set for the different loss and architecture combinations evaluated with the different metrics using the personalized data scheme. Predictions for each trajectory have been accumulated and the metrics are evaluated on the resulting accumulation. 

\begin{table*}[!t]
    \centering
    \begin{tabular}{|c|c|c|c|c|c|}
        \hline
        Network & Param. (M) & $L_2$ (voxel) & CE \\
%        \hline
%         DeepEndoVO \citep{turan_deep_2018} & \textbf{1993.66} &   &    $\pm$  \\
         \hline
         OffsetNet \citep{sganga_deep_2019} & 43.6 & 5.694 $\pm$ 4.634 &  0.994 $\pm$ 0.342 \\
         \hline
         BronchoTrack &  \textbf{14.1}  & \textbf{4.994 $\pm$ 4.442} &  \textbf{0.808 $\pm$ 0.354} \\
         \hline
    \end{tabular}
    \caption{Comparison with state of the art (SoA) method \citep{sganga_deep_2019} for bronchoscopy tracking in terms of position error ($L_2$), angle error, and the number of parameters. Results are computed after accumulating tracking through the whole test paths and only by evaluating last position and orientation, as indicated in Equation~\ref{eq:posesteps}. Values show mean and standard deviation computed among all the paths in the test set.}
    \label{tab:stateoftheart}
\end{table*}

\begin{figure*}[!t]
    \centering
    \begin{subfigure}{0.32\textwidth}
        \centering
        \includegraphics[width=0.9\linewidth]{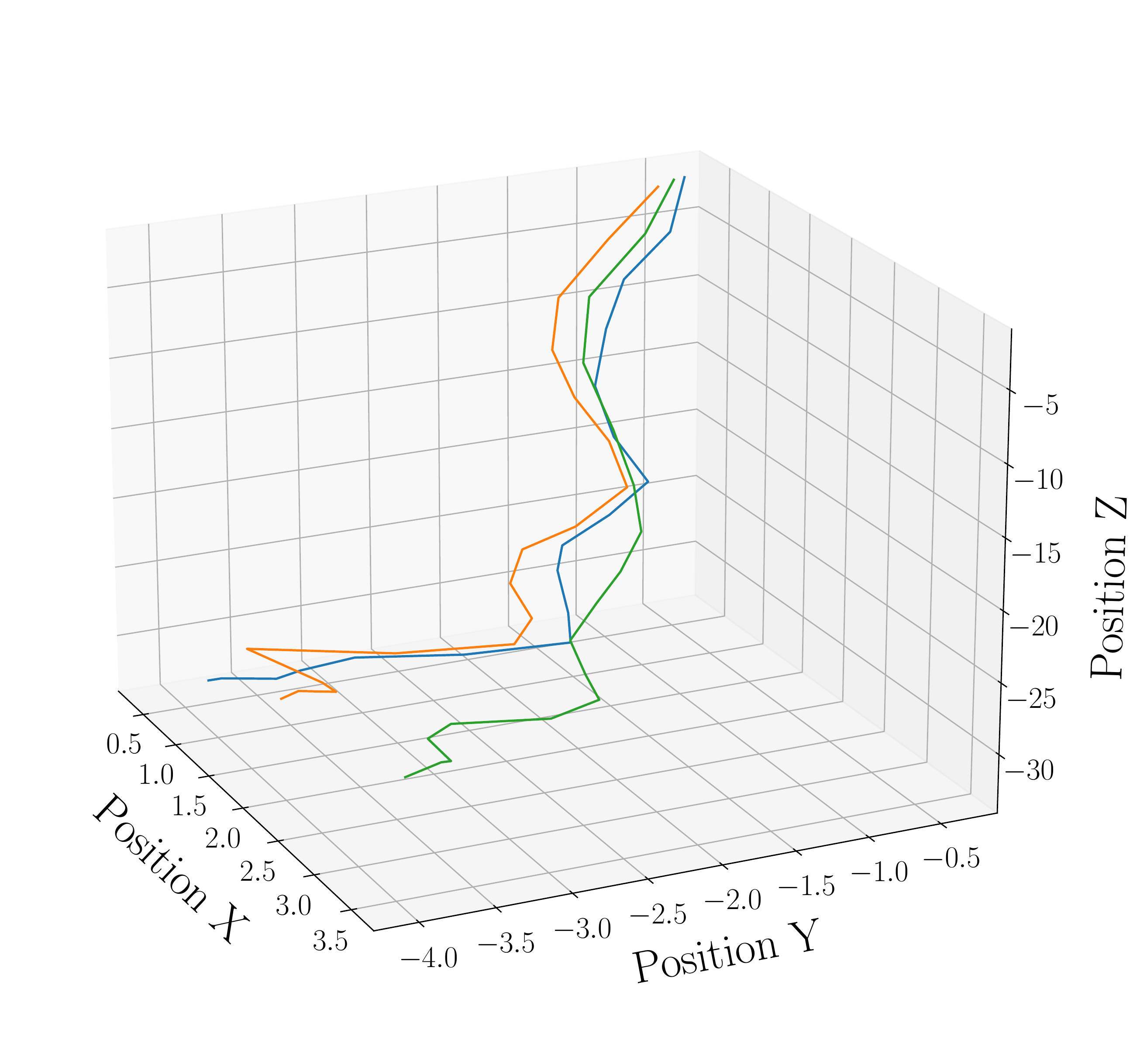}
    \end{subfigure}
    \begin{subfigure}{0.32\textwidth}
        \centering
        \includegraphics[width=0.9\textwidth]{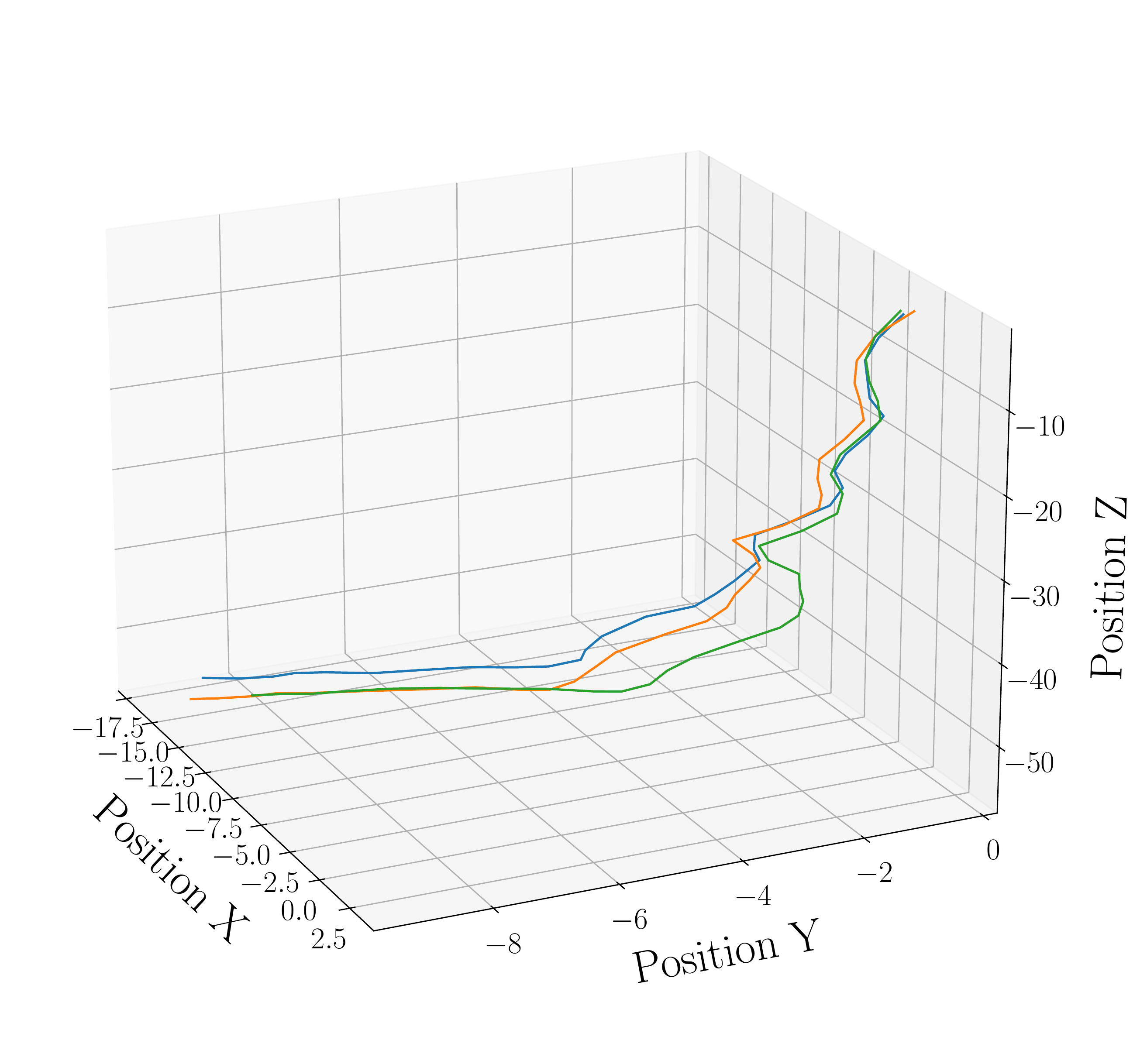}
    \end{subfigure}
    \begin{subfigure}{0.32\textwidth}
        \centering
        \includegraphics[width=0.95\textwidth]{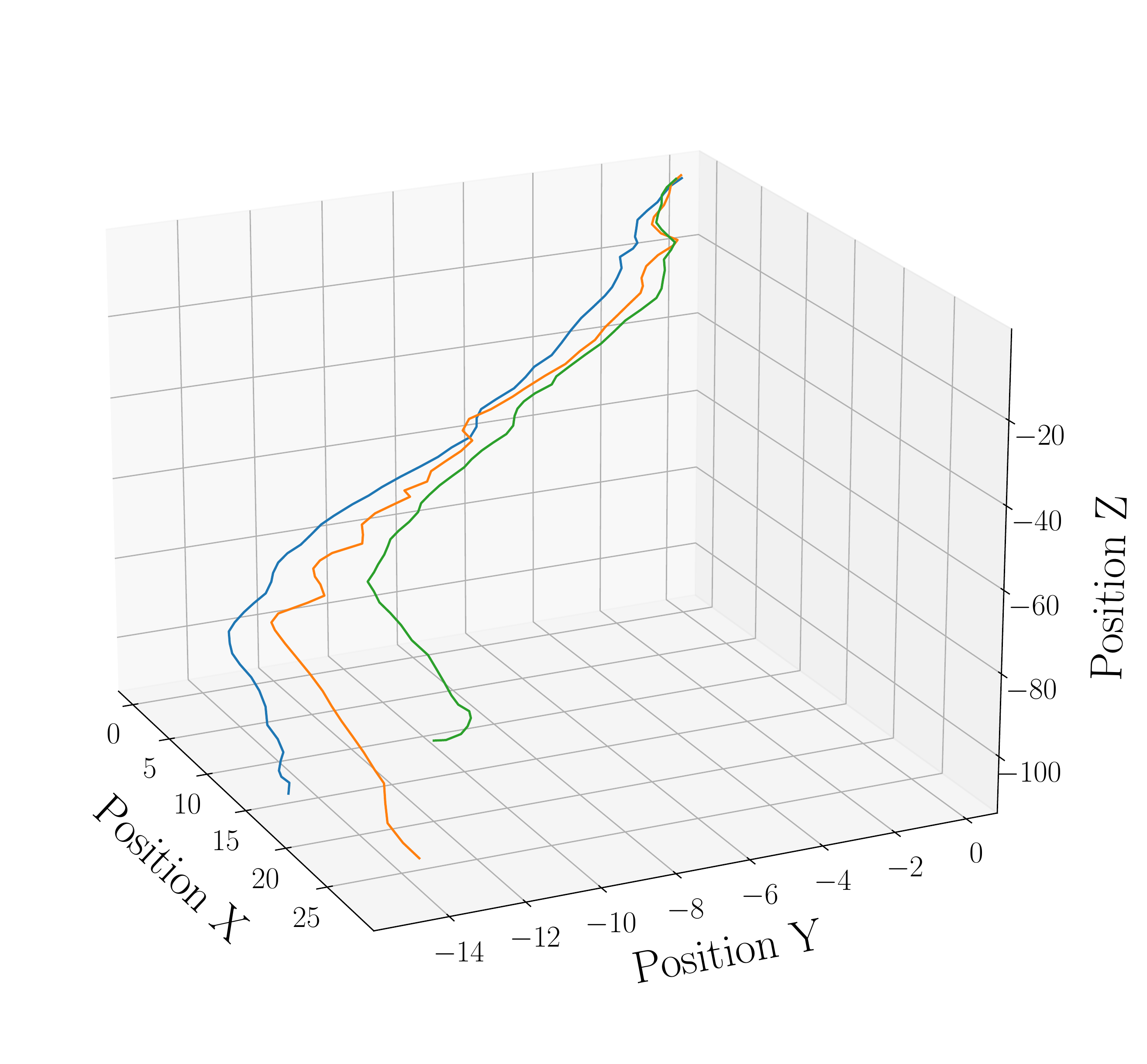}
    \end{subfigure}
    \caption{Sample position tracking comparison with SoA method, OffsetNet (green), our method (orange), and the ground truth (blue). Axis coordinates are defined in voxel units, with the (0, 0, 0) being in the trachea.} %Left, LENS_P20_14_01_2016_INSP_CPAP_bl_0, center, LENS_P21_04_02_2016_INSP_CPAP_ur_0, right, LENS_P30_14_04_2016_INSP_CPAP_br_1. \label{fig:trajectory_comparison}}
    \label{fig:trajectory_comparison}
\end{figure*}

Regardless of the loss, the best values with regards to position error are achieved, with notable difference, by those architectures with temporal information management, indicating the importance of it. Moreover, the combination of ME + CE losses provides the first and second best results, proving its usefulness for better positioning. With regards to rotation error, we can point several interesting points. First, temporal management information structures notably improve CE metric result, with the a great improvement with Convolutional RNN. Second, MSE seems to provide better results in both $L_2$ and CE metrics, and with the exception of DE. And, finally, in the case of DE and $L_2$ temporal information information management architectures seem not to provide any benefits.

Differently to Table~\ref{tab:lossresultsacum}, Table~\ref{tab:lossresults} shows the results for the same configurations but evaluating over image pairs and not whole trajectories. In this case, the loss combination MSE + CE still provides the best position estimation and best orientation differences are provided by the MSE + MSE loss, in concordance with results in Table~~\ref{tab:lossresultsacum} . In this case, however, the 3D convolution architecture shows superior performance in orientation estimation.

In general two conclusions can be obtained from Table~\ref{tab:lossresultsacum} and Table~\ref{tab:lossresults} results. First, temporal management information structures help improve both rotation and position predictions, specially in the case of ConvRNN architecture for position and 3D for orientation. Secondly, the loss combination MSE + CE delivers better results for position prediction, while MSE + MSE improves rotation precision.

\subsection{Model generalization}

In Tables~\ref{tab:archablationsacum} and \ref{tab:archablations} we show results for the cross-subject or population data scheme defined in Section~\ref{subsec:population}. Architecture  and loss combinations are selected from the results in the personalized data scheme evaluation. Table~\ref{tab:archablationsacum} shows results evaluated at the last position and orientation after accumulating trajectory changes, and Table~\ref{tab:archablations} shows results for each image pair.

As seen, with regards to position error, the combination of ConvRNN and MSE + CE loss delivers notably improved results compared to the other two alternatives, showcasing its better generalization capabilities with unseen patients. With regards to orientation errors, the 3D convolution mechanism provides, with difference, best results in orientation, both with metrics $L_2$ and CE. Followed by ConvRNN in the accumulated case and by LSTM in the image-pair case. Hence, it can be stated that convRNN plus MSE + CE as loss is the best option for position prediction, while 3D convolutions plus MSE + MSE is the best one for rotation prediction. 

Importantly, another fact that can be observed is the difference in results between the personalized (Tables~\ref{tab:lossresultsacum} and \ref{tab:lossresults}) and the population data schemes (Tables~\ref{tab:archablationsacum} and \ref{tab:archablations}). In the former, results are much better for both for orientation and position with metric $L_2$, although the effect in CE and DE is smaller. Such point showcases the benefits of the personalized model, and the usefulness of retraining the model with each patient information.

%\subsubsection{Patient Settings}

%We implement the two patient settings described in Section~\ref{subsec:patientsetting}, outer and intra settings, altogether with the loss configuration, $\mathfrak{L_p}_{MSE} + \mathfrak{L_o}_{CE}$. In Figure~\ref{fig:intraouter}, we compare validation performance with regards to position and angle errors of both options. Clearly, intra-patient setting obtains notably better results, both for position and angle results, indicating the benefits of retraining the model each time a new patient enters the pool, albeit its associated costs.

%Moreover, we also examine the effect of increasing the number of training and validation sequences. As seen in Figure~\ref{fig:intraouter}, increasing data helps reduce the error in the intra-patient setting for both the position and angle terms. However, in the case of the outer-patient setting, it only helps reducing the angle error, while the position error is only reduced slightly. The improvements in the intra-patient setting when increasing data show a positive association with our synthetic dataset, since its data production is cheap and unlimited with regards to quantity. 

\subsection{Comparison to State-of-the-art}

Finally, we compare our model with current SoA for bronchoscopy tracking, OffsetNet \citep{sganga_autonomous_2019}, implementing it as described in the original publication. For such comparison we build on our best results in previous sections and choose as loss the combination $\mathfrak{L_p}_{MSE} + \mathfrak{L_o}_{CE}$, and as architecture, the convolutional recurrent network. We call this network BronchoTrack. Both networks are trained and tested in an intra-patient setting, using sequences of 10 steps long for training, and full sequences for testing. 

In Table~\ref{tab:stateoftheart} we show results for the comparison in the test set. As shown our network is better for both position and orientation tracking. Moreover, it has a notably reduced number of parameters: more than 3x times less parameters.

In Figure~\ref{fig:trajectory_comparison} we further compare both models by showing the accumulated position tracking for sample trajectories in the test set. As seen our model is able to closely match the ground truth sequence while OffsetNet deviates more and struggles to follow the ground truth path.

\section{Conclusions}
\label{sec:conclusions}

In the present study we have presented several contributions to bronchoscopy tracking.
We have built a synthetic dataset to allow for fair comparison between methods and be able to train data-hungry models. Two different population approaches for training learning methods have been analyzed, concluding with the benefits of a personalized setting. We also have experimented with the configuration of a neural network model with regards to the loss function and the temporal information management. Finally, when comparing to a SoA method, better results have been obtained while using less parameters. All in all, there are still important next steps to take to improve current results. Such steps could involve the study of transfer learning to real bronchoscopy videos, for example with the use of generative adversarial networks, or the adaptation to the specific conditions of a medical setting in terms of hardware and resources.

\section*{CRediT Authorship Contribution Statement}

\textbf{Juan Borrego-Carazo}: Conceptualization, Formal analysis, Investigation, Methodology, Software, Validation, Visualization, Roles/Writing - original draft, Writing - review $\&$ editing. \textbf{Carles Sánchez}:  Conceptualization, Data curation, Formal analysis, Supervision, Visualization, Roles/Writing - original draft, Writing - review $\&$ editing. \textbf{David Castells-Rufas}: Conceptualization, Methodology, Supervision, Validation,  Writing - review $\&$ editing. \textbf{Jordi Carrabina}: Funding acquisition, Project administration, Resources, Supervision, Writing - review $\&$ editing. \textbf{Débora Gil}: Conceptualization, Formal analysis, Funding acquisition, Investigation, Methodology, Project administration, Resources, Supervision, Validation, Roles/Writing - original draft, Writing - review $\&$ editing.

\section*{Declaration of Competing Interest}

The authors declare that they have no known competing financial
interests or personal relationships that could have appeared to influence the work reported in this paper.

\section*{Acknowledgements}
Funded by Ministerio de Ciencia e Innovación (MCI), Agencia Estatal de Investigación (AEI) and Fondo Europeo de Desarrollo Regional (FEDER), RTI2018-095209-B-C21, RTI2018-095209-B-C22  (MCI/AEI/FEDER, UE), Generalitat de
Catalunya, 2017-SGR-1624, CERCA-Programme and the Catalan Government
industrial Ph.D. program under grant 2018-DI-30. DGil is Serra Hunter.

\bibliography{broncho}

\end{document}